\documentclass[10pt,conference]{IEEEtran}
\IEEEoverridecommandlockouts
\usepackage{cite}
\usepackage{amsmath,amssymb,amsfonts}
\usepackage{algorithmic}
\usepackage{graphicx}
\usepackage{textcomp}
\usepackage{adjustbox}
\usepackage{multirow}
\usepackage{xcolor}
\usepackage{subcaption}
\def\BibTeX{{\rm B\kern-.05em{\sc i\kern-.025em b}\kern-.08em
    T\kern-.1667em\lower.7ex\hbox{E}\kern-.125emX}}
\begin{document}

\title{Online Detection of Water Contamination Under Concept Drift\\
\thanks{This paper was supported by the European Research Council (ERC) under grant agreement No 951424 (Water-Futures), the European Union’s Horizon 2020
research and innovation programme under grant agreement No 739551 (KIOS CoE), and the Republic of Cyprus through the Deputy Ministry of Research, Innovation and Digital Policy.}% <-this % stops a space
}

\author{
	\IEEEauthorblockN{
		Jin Li\textsuperscript{1, 2},
		Kleanthis Malialis\textsuperscript{1},
        Stelios G. Vrachimis\textsuperscript{1},
		Marios M. Polycarpou\textsuperscript{1, 2}
	}
 
	\IEEEauthorblockA{
		\textsuperscript{1} \textit{KIOS Research and Innovation Center of Excellence}\\
		\textsuperscript{2} \textit{Department of Electrical and Computer Engineering}\\
        University of Cyprus, Nicosia, Cyprus\\
		\{li.jin, malialis.kleanthis, vrachimis.stelios, mpolycar\}@ucy.ac.cy
		\\ORCID: \{0000-0002-3534-524X, 0000-0003-3432-7434, 0000-0001-8862-5205, 0000-0001-6495-9171\}
	}
}

\maketitle

\begin{abstract}
Water Distribution Networks (WDNs) are vital infrastructures, and contamination poses serious public health risks. Harmful substances can interact with disinfectants like chlorine, making chlorine monitoring essential for detecting contaminants. However, chlorine sensors often become unreliable and require frequent calibration. This study introduces the Dual-Threshold Anomaly and Drift Detection (AD\&DD) method, an unsupervised approach combining a dual-threshold drift detection mechanism with an LSTM-based Variational Autoencoder (LSTM-VAE) for real-time contamination detection. Tested on two realistic WDNs, AD\&DD effectively identifies anomalies with sensor offsets as concept drift, and outperforms other methods. A proposed decentralized architecture enables accurate contamination detection and localization by deploying AD\&DD on selected nodes.

\end{abstract}


\section{Introduction}

Water Distribution Networks (WDNs) are crucial for community well-being and economic growth, requiring robust monitoring to address challenges like contamination detection. For instance, Milwaukee, USA, experienced one of the largest U.S. waterborne outbreaks when Cryptosporidium contaminated the water supply, affecting over 400,000 residents \cite{mac1994massive}. Timely detection is essential to protect public health, ensure safe drinking water, and mitigate risks.

Chlorine is injected into WDNs for disinfection, with concentrations maintained within specific bounds for effective water quality management \cite{constans2003simulation}. Accurate monitoring is crucial under varying conditions, as chlorine levels fluctuate due to factors like chemical reactions or malicious attacks, serving as potential contamination indicators \cite{vrachimis2024disinfection}. Manual sampling methods are ineffective during contamination events due to delays, while online monitoring ensures prompt threat detection and response.

With advancements in environmental sensing networks, data-driven analytics are increasingly used to study and predict water quality \cite{kang2017data}. However, real-world data often exhibits non-stationary behavior, or concept drift, due to factors like measurement offsets, power failures, and hydraulic changes. Distinguishing concept drift from continuous anomalies, such as persistent changes in chlorine concentration, becomes challenging when both occur simultaneously.

This study addresses contamination detection and localization in non-stationary WDN environments. We propose the Dual-Threshold Anomaly and Drift Detection (AD\&DD) method and a decentralized architecture for localization. The key contributions are:

\begin{enumerate}
   \item We evaluate two realistic water networks with strategically placed sensors, classifying arsenite contamination as anomalies and sensor offsets as concept drift. In the decentralized architecture, each sensor uses AD\&DD for local anomaly detection, enabling fault localization with initial flow direction knowledge.

    \item AD\&DD employs an LSTM-VAE-based online learning algorithm with a dual-threshold mechanism for detecting both concept drift and anomalies without supervision. Drift detection occurs in the latent space. Comparative analysis shows that our method outperforms several state-of-the-art techniques.

\end{enumerate}

The paper is organized as follows: Section~\ref{sec:related} reviews related work. Section~\ref{sec:method} details the proposed detection method and architecture. Sections~\ref{sec:exp_setup} and \ref{sec:exp_results} cover the experimental setup and results. Finally, Section~\ref{sec:conclusion} concludes with a summary and future work. Code and data will be available upon acceptance.

\section{Related Work}\label{sec:related}

\subsection{Concept drift adaptation} 
Data nonstationarity, often due to concept drift (a change in the underlying probability distribution), is a challenge in streaming applications. Methods to address drift are classified as passive or active\cite{ditzler2015learning}. \textbf{Passive methods} adapt incrementally without full re-training\cite{malialis2020online,hulten2001mining,liang2006fast,widmer1996learning}. \textbf{Active methods} detect changes in the data distribution to trigger adaptation. \textbf{Hybrid methods} combine the strengths of both approaches\cite{malialis2022hybrid}.

Requiring fully labeled data can be unrealistic in some real-world scenarios. To address this issue, the research community has explored alternative learning paradigms. In \textbf{active learning}\cite{ malialis2022nonstationary}, strategies are employed to intelligently determine when to query a human expert for ground truth information, such as class labels, for selected examples. \textbf{Unsupervised learning}, where no labeled data is required, autoencoders have emerged as effective drift detectors. For instance, \cite{jaworski2020concept} introduces an autoencoder-based approach that monitors two distinct cost functions—cross-entropy and reconstruction error—to detect concept drift. The variation in these cost functions serves as an indicator for concept drift detection. Another work \cite{li2023autoencoder} also employs autoencoder and leverages the advantages of both incremental learning and drift detection based on Mann-Whitney U Test. In this paper, we focus on unsupervised learning methods and AEs.

\subsection{Contamination diagnosis}\label{sec:conta}

\subsubsection{Contamination detection} Machine learning techniques classify water quality time series as normal or anomalous, as reviewed in \cite{kang2017data}.  Deep Belief Networks (DBNs), an unsupervised learning method, effectively reconstruct inputs probabilistically. In \cite{solanki2015predictive}, DBNs are used to analyze and predict water chemical features, showing superior accuracy over traditional supervised learning methods. Threshold methods, like Isolation Forest (iForest) \cite{liu2020integrated}, set upper and lower limits to classify data. Among density-based algorithms, the Local Outlier Factor (LOF) \cite{breunig2000lof} uses reachability distance to detect outliers, while iForest detects anomalies by analyzing path lengths in an ensemble of trees. VAE4AS \cite{li2024unsupervised} introduces dual concept drift detection, distinguishing between abnormal sequences and drift, making it suitable for contamination detection under concept drift.

\subsubsection{Contamination localization}

Several approaches focus on contamination localization. \cite{grbvcic2020machine} combines Artificial Neural Networks (ANNs) for pollution classification and Random Forests for regression analysis, using Monte Carlo simulations to identify contamination sources, though it remains offline in stationary environments. \cite{costa2013localization} proposes a method for locating contamination sources in DWDS, but it assumes known flow directions and does not address concept drift. In \cite{sankary2019bayesian}, an inline mobile sensor localizes contamination based on Bayesian updates of intrusion probabilities.

These methods highlight a research gap in real-time online contamination localization under concept drift. To address this, we propose a cost-effective approach using fixed sensors, capable of detecting anomalies and localizing contamination in real-time with only initial flow direction data.

% \begin{figure*}[!h]
% \centering
% \begin{subfigure}{0.5\textwidth}
%   \centering
%   \includegraphics[width=0.63\columnwidth]{image/arch_decen_new.png}
%   \caption{Decentralized Architecture}
%   \label{fig:decen}
% \end{subfigure}%
% \begin{subfigure}{0.5\textwidth}
%   \centering
%   \includegraphics[width=0.85\columnwidth]{image/arch_cen_new.png}
%   \caption{Centralized Architecture}
%   \label{fig:cen}
% \end{subfigure}

% \caption{Illustration of detection architecture.}
% \label{fig:both_arch}
% \end{figure*}

\section{Online Water Contamination Detection and Localization}\label{sec:method}
\subsection{Overview}

\textbf{Problem formulation}. A WDN is defined as a network, with nodes representing junctions and edges denoting pipes. The task is to identify any harmful substance which might have been injected accidentally (e.g., during a leakage) or maliciously. We identify it at points where it reacts with chlorine, which is already in the water for disinfection purposes. The concept drift considered here is the measurement offset of sensors. The chlorine sensors are installed on the nodes to measure the concentration level. These sensors are represented as $S_{1}, S_{2}, ..., S_{K}$. At each time $t$, the univariate time series of $S_{i}$ is $x^t_{i}\in R$, where $i\in(1, 2,..., K)$.

\begin{figure}[!t]
	\centering
	\includegraphics[scale=0.26]{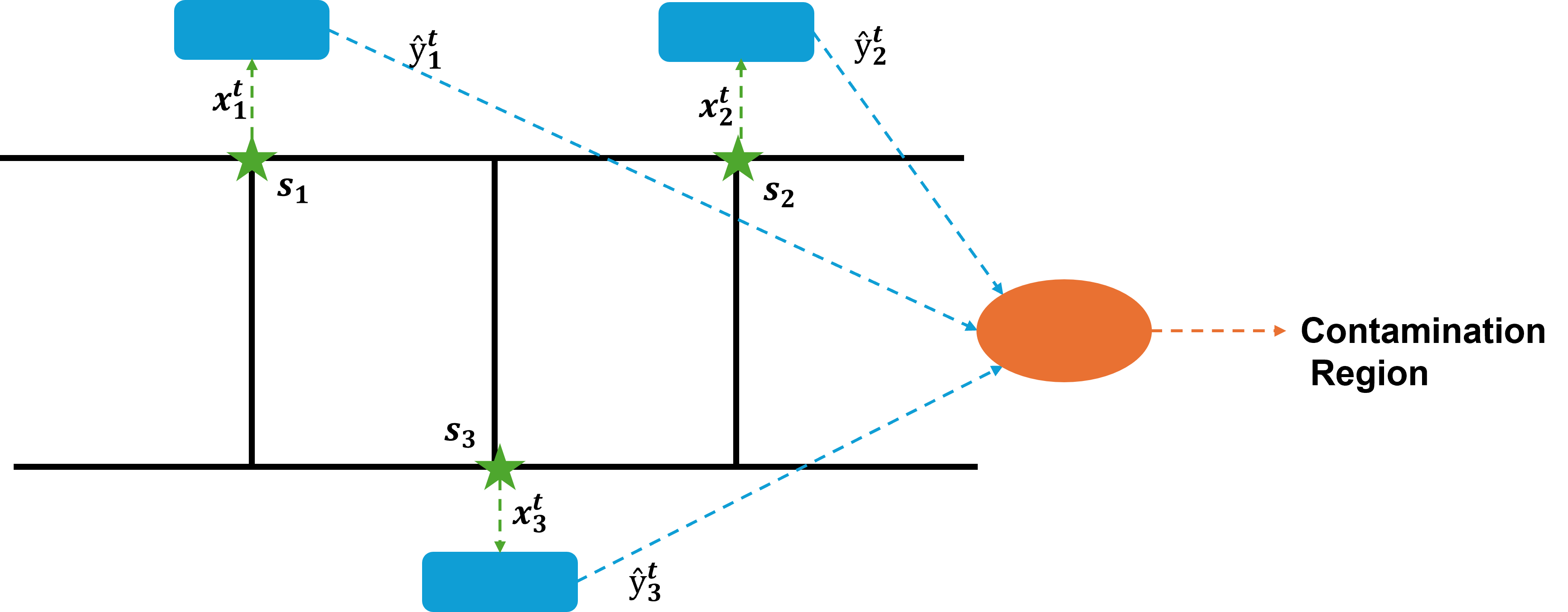}
	\caption{Illustration of decentralized architecture.}
	\label{fig:arch}
\end{figure}

\begin{figure}[!t]
	\centering
	\includegraphics[scale=0.26]{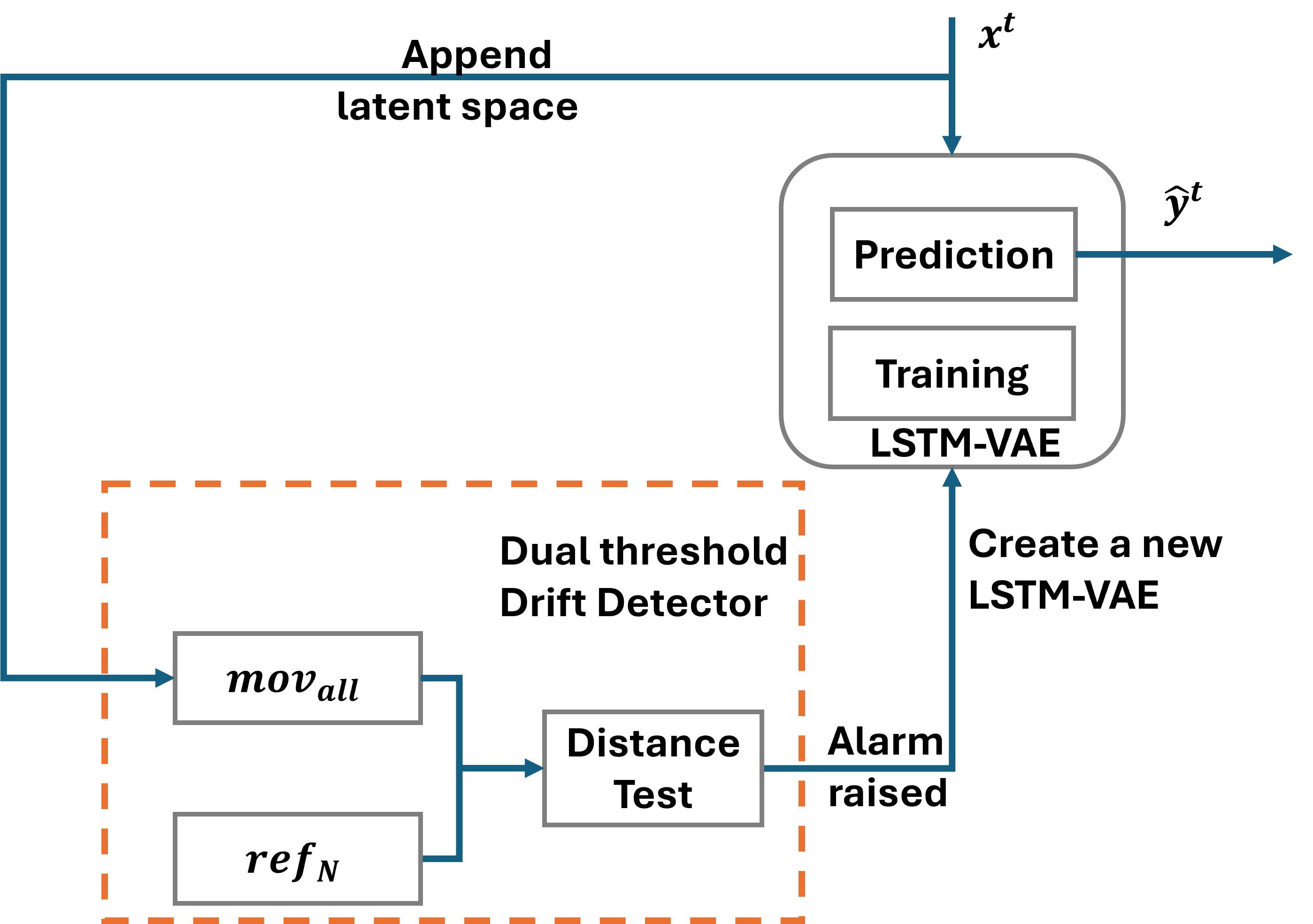}
	\caption{Overview of AD\&DD.}
	\label{fig:algo}
\end{figure}

\textbf{Sensor placement}. In practice, sensor installation is optimized by selecting a subset of nodes from feasible locations to minimize objectives like risk, considering impact metrics such as the number of people infected \cite{ostfeld2008battle}. This study uses the strategy from \cite{eliades2014sensor} with a multi-objective evolutionary algorithm to minimize infected populations. Additional methods are discussed in \cite{hu2018survey}.

\textbf{Proposed Architecture}. This study proposes a decentralized architecture where each sensor hosts a detector, sending results to a monitoring center for real-time contamination localization. Fig.~\ref{fig:arch} illustrates the architecture, with green dashed lines for sensor measurements, blue dashed lines for AD\&DD predictions, and the orange section highlighting contamination localization. A detailed technical description follows.

\textbf{Proposed algorithm}. The unsupervised AD\&DD algorithm detects anomalies and localizes them. As shown in Fig.~\ref{fig:algo}, the system observes the value at time $t$, denoted as $x^t$. In the decentralized architecture, the observation is $x^t_i \in R$ from sensor $s_{i}$. The output $\hat{y}^t \in \{0, 1\}$ indicates anomaly (1) or normal conditions (0).

Autoencoders detect anomalies by reconstructing normal data with minimal error. To capture temporal dependencies, we integrate a Variational Autoencoder (VAE) with Long Short-Term Memory (LSTM), addressing the vanishing gradient problem. The VAE learns a distribution $q(z|x)$ with a multivariate Gaussian assumption, and regularization is applied via Kullback-Leibler (KL) divergence. The total VAE loss combines reconstruction loss and KL divergence, with $\beta \geq 0$ adjusting KL's weight. Our algorithm also includes a drift detection component to adapt $l_{VAE}$.

\begin{equation}\label{eq:vae}
l_{V A E}(x, \hat{x})=l_{A E}(x, \hat{x})+ \beta * l_{K L}(x)
\end{equation}

\subsection{Detailed Description}\label{sec:decen}

\textbf{Deployment}. In the decentralized architecture, each sensor has its own AD\&DD mechanism for localized monitoring. Let $\hat{y}^t_1, \hat{y}^t_2, \dots, \hat{y}^t_K$ represent anomaly detection results at sensors $S_1, S_2, \dots, S_K$ at time $t$, where $\hat{y}^t_i \in \{0, 1\}$. This enables localized anomaly detection, aiding fault localization when combined with flow direction. A central monitoring center can analyze real-time results for network-wide fault localization.

\textbf{Memory}. Each AD\&DD has three windows: $ref_{N}$, $mov_{all}$ and $mov_{retrain}$. Therefore, the number of windows is $3*K$. $ref_{N}$ stores the encodings of normal instances and $mov_{all}$ stores the encodings of the most recent instances. Both $ref_{N}$ and $mov_{all}$ have the same size as $W_{drift}$. $mov_{retrain}$ stores the instances of size $W_{retrain}$ once drift alarm is raised. 

The following descriptions are for individual AD\&DD.

\textbf{AD (Anomaly Detection)}. An adaptive threshold $\theta^t$ is set based on the maximum training loss, with anomalies detected when the cumulative loss exceeds this threshold. The initial threshold is calculated using offline data. At retraining time $t$, the loss for all elements in the window $mov_{retrain}$ is calculated: $L^t = \{l_{VAE}(x^i, \hat{x}^i)\}^t_{i=t-W_{retrain}+1}$. The threshold at training time $t$ is set to avoid false drift alarms:
\begin{equation}\label{eq:adtthreshold}
\theta^t = max(L^t)+std(L^t)
\end{equation}

\textbf{DD (Drift Detection)}. Drift is detected by calculating the Euclidean distance between the reference window $ref_{N}$ and the most recent data window $mov_{all}$, as shown in Eq.~(\ref{eq:ed}). Dual distance thresholds help distinguish between minor sensor drifts and significant deviations indicating contamination, as shown in Eq.~(\ref{eq:ed_flag}). If a drift is detected, the model is retrained, and the reference window is updated with data in $mov_{retrain}$.

\begin{equation}\label{eq:ed}
\text\ Dis(ref_{N}, mov_{all})=\sqrt{\sum_{i=1}^n \sum_{j=1}^m\left(ref_{Nij}-mov_{all i j}\right)^2}
\end{equation}

\begin{equation}\label{eq:ed_flag}
\begin{aligned}
{alarm} = True \ &  if \ thre_{upp} > Dis(ref_{N}, mov_{all}) > thre_{low}\\
\end{aligned}
\end{equation}

% \textbf{Localisation}. For sensors that detect an anomaly at time $t$ (i.e., \( y_i^t = 1 \)), their positions are correlated with the known flow direction, and the sensor located at the most upstream position is identified as the fault source. Specifically, define \( \mathcal{S}^t = \{S_i^t \mid \hat{y}_i^t = 1\} \) as the set of sensors that detect an anomaly at time $t$. Within the set \( \mathcal{S}^t \), the most upstream sensor \( \text{S}_{i^*}^t \) is selected as the fault source based on the known flow direction, i.e.,
% \begin{equation}\label{eq:fl}
% i^* = \arg\min_{i \in \mathcal{S}^t} \text{Position}(i)
% \end{equation}
% where \( \text{Position}(i) \) represents the position index of sensor $S_{i}$ in the water system, with smaller values indicating positions closer to the upstream.

\textbf{Localisation}. The fault is assumed to lie in the intersection of the upstream region of the sensors that detect an anomaly at time $t$ (i.e., \( y_i^t = 1 \)) and the downstream region of the sensors that do not detect an anomaly. Specifically, define \( \mathcal{S}_1^t = \{S_i^t \mid \hat{y}_i^t = 1\} \) as the set of sensors that detect an anomaly at time $t$, and \( \mathcal{S}_0^t = \{S_i^t \mid \hat{y}_i^t = 0\} \) as the set of sensors that do not detect an anomaly at time $t$. The contamination region is the intersection of the upstream of \( \mathcal{S}_1^t \) and the downstream of \( \mathcal{S}_0^t \):
\begin{equation}\label{eq:loc}
\text{Contamination Region} = \text{Upstream}(\mathcal{S}_1^t) \cap \text{Downstream}(\mathcal{S}_0^t)
\end{equation}

\section{Experimental Setup}\label{sec:exp_setup}
\subsection{Data Generation}\label{sec:data_gen}
\begin{figure}[!t]
\centering
\begin{subfigure}[b]{0.5\textwidth}
  \centering
  \includegraphics[width=0.65\linewidth]{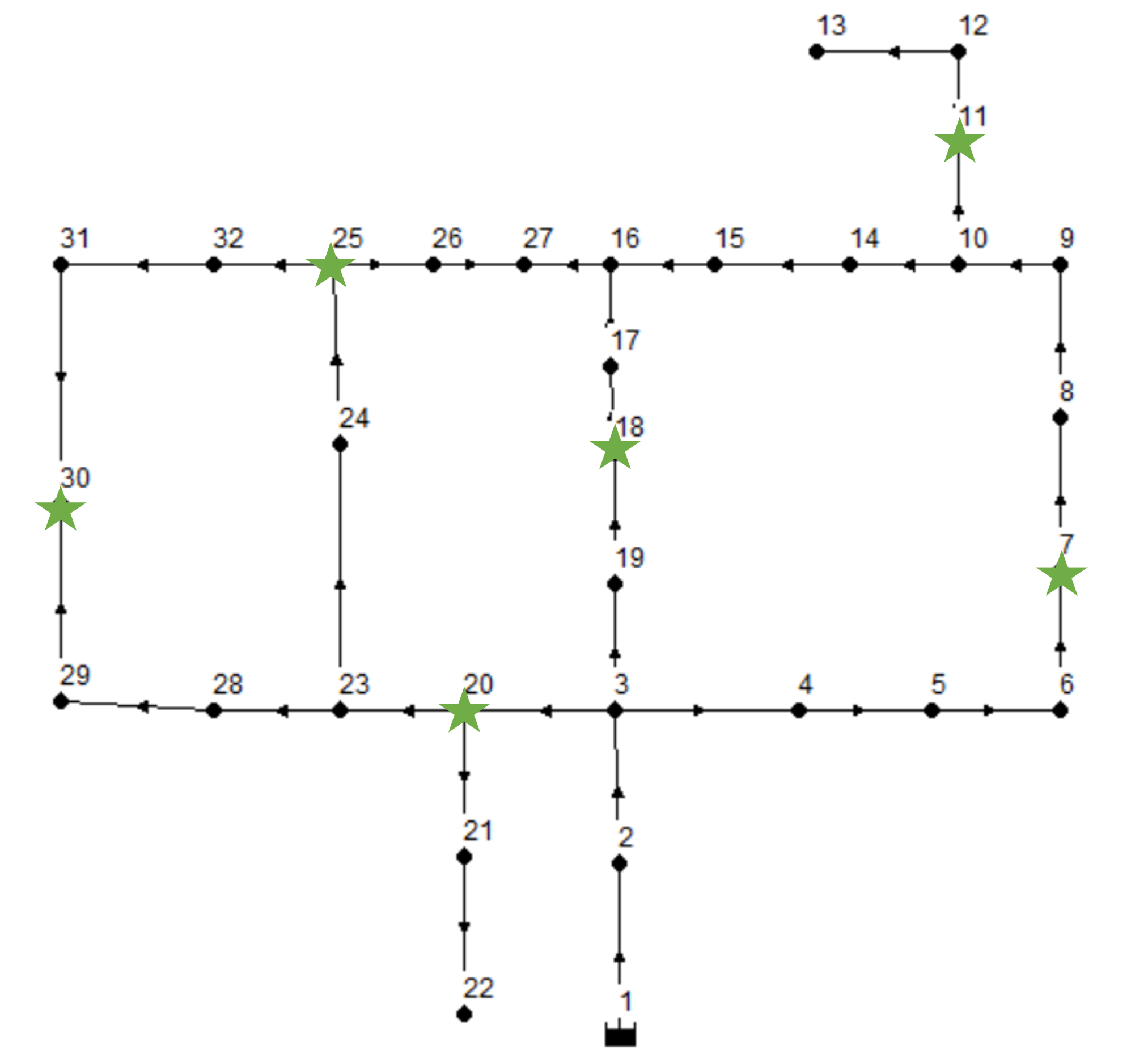}
  \caption{Hanoi network}
  \label{fig:sen_hanoi}
\end{subfigure}%
\hfill
\begin{subfigure}[b]{0.5\textwidth}
  \centering
  \includegraphics[width=0.7\linewidth]{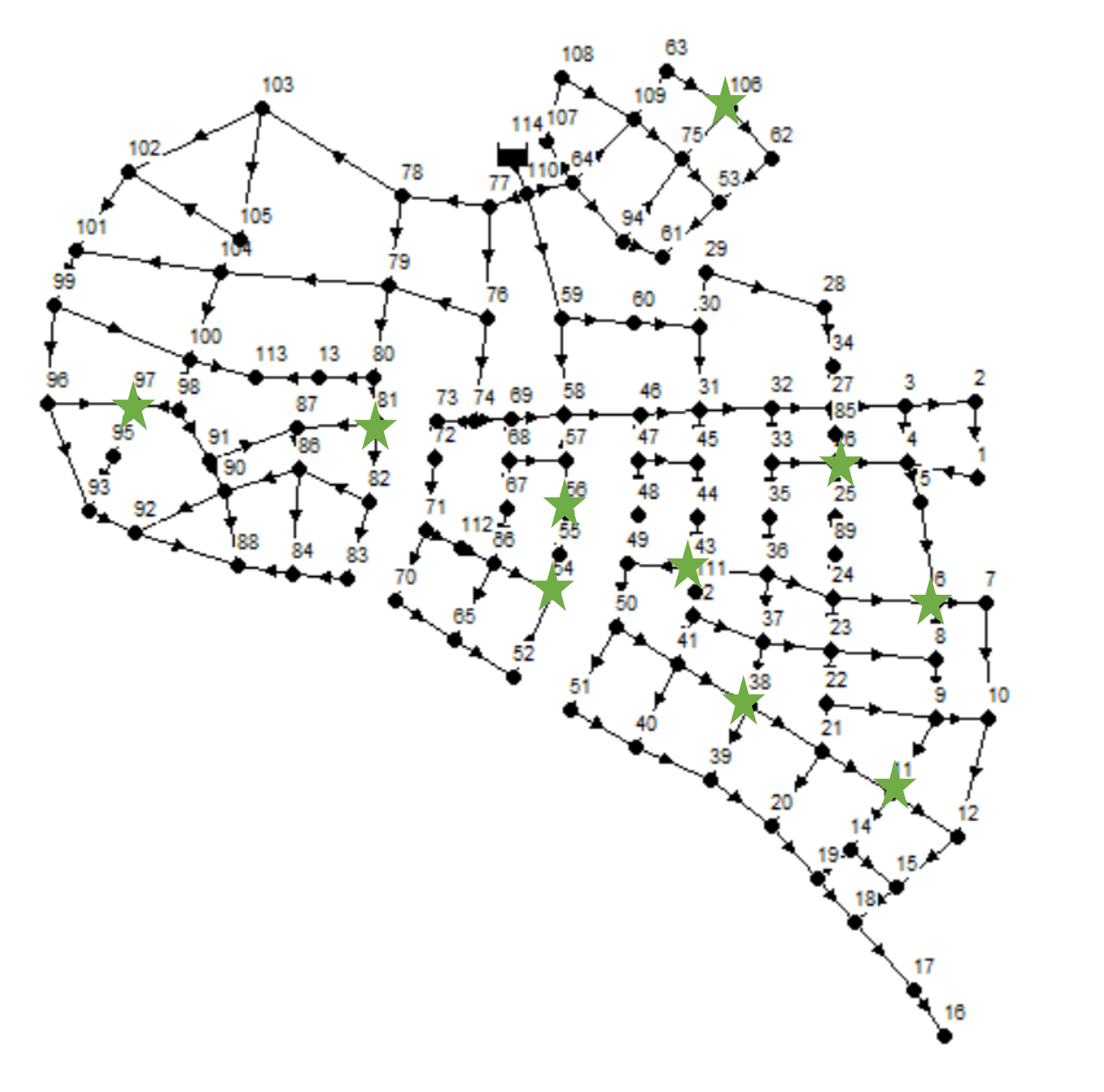}
  \caption{ZJ network}
  \label{fig:sen_ZJ}
\end{subfigure}

\caption{Illustration of network topology and sensor placement.}
\label{fig:sensor_placement}
\end{figure}

\noindent\textbf{Hanoi and ZJ networks.} The Hanoi network has 32 nodes, 34 pipes, and a reservoir \cite{fujiwara1990two}. The Zhi Jiang (ZJ) network features 164 pipes, 113 demand nodes, and 50 primary loops, with a fixed head reservoir \cite{zheng2011combined}. Fig.~\ref{fig:sensor_placement} shows both networks, including flow directions and sensor placements.

\noindent\textbf{Scenarios generation.} 
We use EPyT-Flow \cite{artelt2024toolbox} to model scenarios in WDNs, creating three scenarios each for the Hanoi and ZJ networks based on demand patterns from \cite{vrachimis2018leakdb}. Chlorine (0.7 mg/L) is continuously injected for disinfection, with arsenite (0.8-1 mg/L) introduced in fault scenarios. Sensor offsets are set to 0.98 of the true value. Six months of historical normal data are used for pretraining, followed by six months with faults and drifts for online training. STL \cite{cleveland1990stl} is applied to adjust data, removing trend and residual components, with a one-week period. Anomaly periods are 960-1440 and 5760-6240, as shown in Table~\ref{tab:sce}.

\begin{table}[!t]
\caption{Scenarios description of Hanoi and ZJ}\label{tab:sce}
\centering
\begin{adjustbox}{width=0.5\textwidth}
\begin{tabular}{|cc|cccccc|}
\hline
\multicolumn{2}{|c|}{Network}                                                                                                             & \multicolumn{3}{c|}{Hanoi}                                                                                                                                                                                                                                              & \multicolumn{3}{c|}{ZJ}                                                                                                                                                                                                                            \\ \hline
\multicolumn{2}{|c|}{Scenarios}                                                                                                           & \multicolumn{1}{c|}{Sce.1}                                                            & \multicolumn{1}{c|}{Sce.2}                                                             & \multicolumn{1}{c|}{Sce.3}                                                             & \multicolumn{1}{c|}{Sce.1}                                                             & \multicolumn{1}{c|}{Sce.2}                                                             & Sce.3                                                            \\ \hline
\multicolumn{1}{|c|}{\multirow{2}{*}{\begin{tabular}[c]{@{}c@{}}Anomalies\\ (Contamination)\end{tabular}}}                 &Location  & \multicolumn{1}{c|}{N5}                                                               & \multicolumn{1}{c|}{N19}                                                                & \multicolumn{1}{c|}{N8}                                                                & \multicolumn{1}{c|}{N4}                                                                & \multicolumn{1}{c|}{N21}                                                               & N33                                                              \\ \cline{2-8} 
\multicolumn{1}{|c|}{}                                                                                                   & Period        & \multicolumn{1}{c|}{\begin{tabular}[c]{@{}c@{}}(960:1440)\\ (5760:6240)\end{tabular}} & \multicolumn{1}{c|}{\begin{tabular}[c]{@{}c@{}}(1440:1920)\\ (5280:5760)\end{tabular}} & \multicolumn{1}{c|}{\begin{tabular}[c]{@{}c@{}}(1200:1680)\\ (5520:6000)\end{tabular}} & \multicolumn{1}{c|}{\begin{tabular}[c]{@{}c@{}}(1440:1920)\\ (5280:5760)\end{tabular}} & \multicolumn{1}{c|}{\begin{tabular}[c]{@{}c@{}}(1200:1680)\\ (5520:6000)\end{tabular}} & \begin{tabular}[c]{@{}c@{}}(960:1440)\\ (5760:6240)\end{tabular} \\ \hline
\multicolumn{1}{|c|}{\multirow{2}{*}{\begin{tabular}[c]{@{}c@{}}Concept Drift\\ (Sensor Offset)\end{tabular}}} & Location & \multicolumn{1}{c|}{N11; N7}                                                          & \multicolumn{1}{c|}{N18; N30}                                                          & \multicolumn{1}{c|}{N11; N18}                                                          & \multicolumn{1}{c|}{N11; N26}                                                          & \multicolumn{1}{c|}{N6; N11}                                                           & N38; N43                                                         \\ \cline{2-8} 
\multicolumn{1}{|c|}{}                                                                                                   & Period         & \multicolumn{6}{c|}{[4000:8640]}                                                                                                                                                                                                                                                                                                                                                                                                                                                                                             \\ \hline
\end{tabular}
\end{adjustbox}
\end{table}

\subsection{Compared methods}
\textbf{iForest++}\cite{liu2020integrated}: An advanced tree-based method for anomaly detection (Sec.~\ref{sec:conta}), adapted to concept drift using incremental learning ('++').

\textbf{LOF++}\cite{breunig2000lof}: Similar to iForest++, we use LOF as the classifier, combined with incremental learning for drift adaptation. The description can be found in Sec.~\ref{sec:conta}.

\textbf{VAE4AS}\cite{li2024unsupervised}: As detailed in Sec.~\ref{sec:conta}, VAE parameters align with AD\&DD, and for drift detection, we use the same statistical test parameters and threshold from original paper.

\textbf{AD\&DD}: The proposed method as described in Sec.~\ref{sec:method}.

\subsection{Performance metrics and evaluation method}
To evaluate experiment results, we use the geometric mean (G-mean) as the performance metric for anomaly detection effectiveness at each node. G-mean is robust to class imbalance \cite{sun2006boosting} and is defined as $G\text{-}mean = \sqrt{R^+ \times R^-}$, where $R^+ = TP / P$ is the positive class recall, and $R^- = TN / N$ is the negative class recall. TP, P, TN, and N represent true positives, total positives, true negatives, and total negatives, respectively. G-mean is insensitive to class imbalance and achieves high values when all recalls are high and their differences are minimal. Prequential evaluation with a 0.99 fading factor converges to the Bayes error on stationary data without a holdout set \cite{gama2013evaluating}. We plot the prequential G-mean at each time step, averaged over 10 repetitions with standard error bars.

\section{Experimental Results}\label{sec:exp_results}
In the following experiments, the hyper-parameters for LSTM-VAE are as follows: learning rate =0.0001, mini-batch size = 64, weight initializer = He Normal; Optimizer = Adam, Hidden activation = Leaky ReLu, Num. of epochs = 100, $\beta$ = 1.0, time step = 10, output activation = Sigmoid, Loss function = Square error, Hidden layers = [2]. $mov_{retrain}$, is set to 500 and $W_{drift}=200$. Optimal thresholds $thre_{low}$ and $thre_{upp}$ for each scenario are determined through experimentation, and the specific values will be provided in the code. 

In the following, we will first analyze the localization capability of the decentralized architecture, and then evaluate AD\&DD's detection performance by comparing it with other approaches.

\begin{table}[!t]
\caption{Decentralized architecture localization performance}\label{tab:loc}
\begin{tabular}{|c|c|c|c|c|}
\hline
         
                                                                                  & \begin{tabular}[c]{@{}c@{}}Hanoi \\ Sce.1\end{tabular} & \begin{tabular}[c]{@{}c@{}}Hanoi\\  Sce.3\end{tabular}  & \begin{tabular}[c]{@{}c@{}}ZJ\\  Sce.2\end{tabular}             & \begin{tabular}[c]{@{}c@{}}ZJ \\ Sce.3\end{tabular}   \\ \hline
\begin{tabular}[c]{@{}c@{}}\# localized nodes\\ during contamination\end{tabular} & \begin{tabular}[c]{@{}c@{}} N(4-6)\\ 3 nodes\end{tabular} & \begin{tabular}[c]{@{}c@{}}N(8-10)\\ 3 nodes\end{tabular} & \begin{tabular}[c]{@{}c@{}} N(8, 9, 21-23)\\ 5 nodes\end{tabular} & \begin{tabular}[c]{@{}c@{}}N(32, 33)\\ 2 nodes\end{tabular} \\ \hline
\# false positive                                                                 & 156                                                    & 266                                                     & 122                                                             & 158                                                   \\ \hline
\# false negative                                                                 & 7                                                      & 15                                                      & 34                                                              & 10                                                    \\ \hline
\end{tabular}
\end{table}

\begin{figure}[!t]
  \centering
 \begin{subfigure}{.25\textwidth}
  \centering
  \includegraphics[width=1.0\columnwidth]{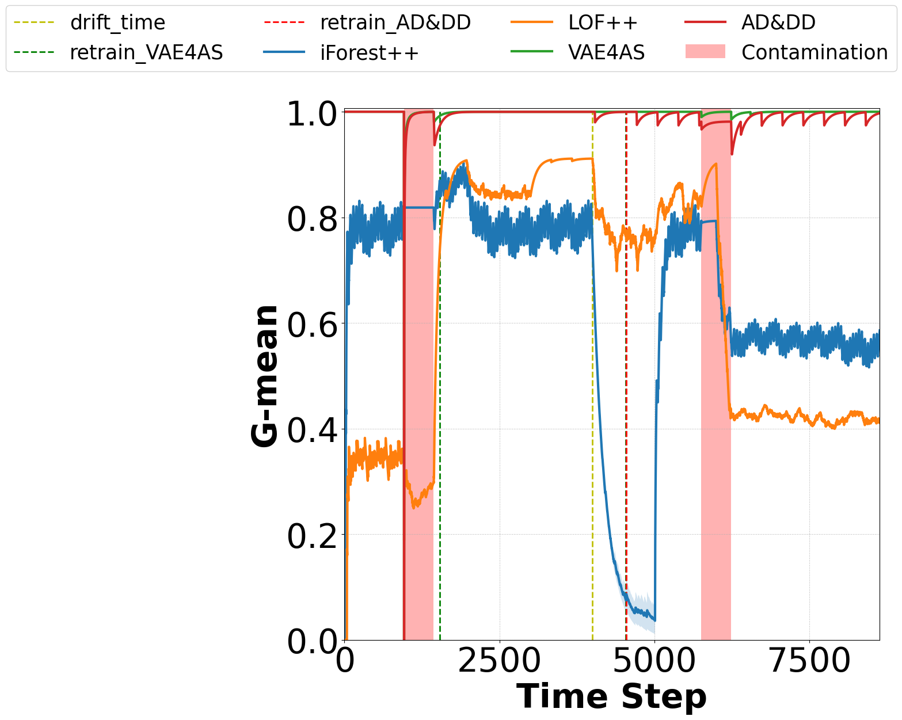} 
  \caption{Hanoi\_Sce.1\_N7}
  \label{fig:Hanoi1_7_com} 
 \end{subfigure}%
 \begin{subfigure}{.25\textwidth}
  \centering
  \includegraphics[width=1.0\columnwidth]{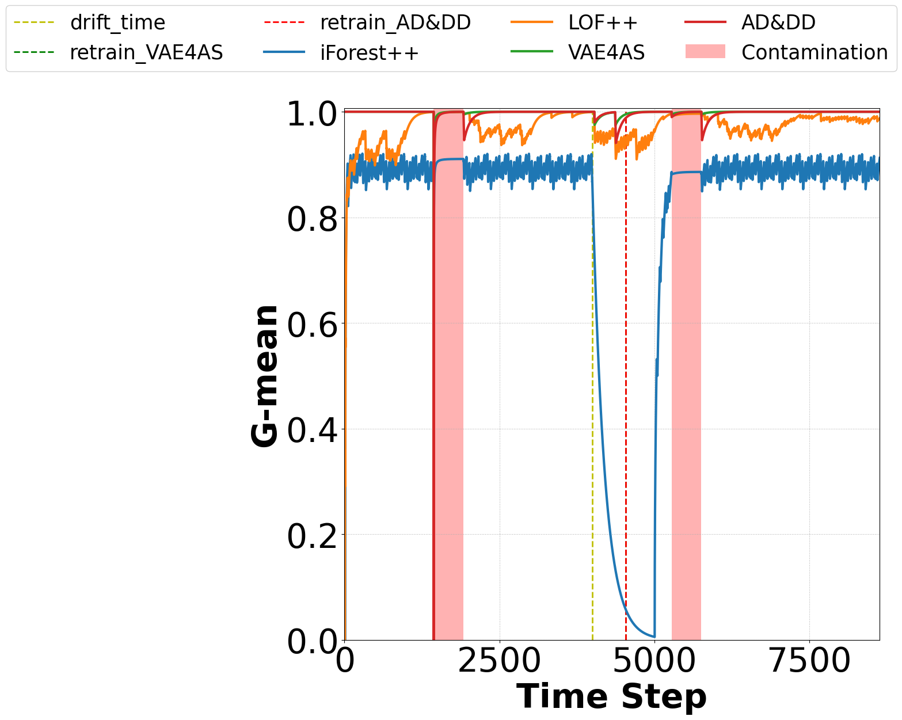} 
  \caption{Hanoi\_Sce.2\_N18}
\label{fig:Hanoi2_18_com}
 \end{subfigure}%

 \begin{subfigure}{.25\textwidth}
  \centering
  \includegraphics[width=1.05\columnwidth]{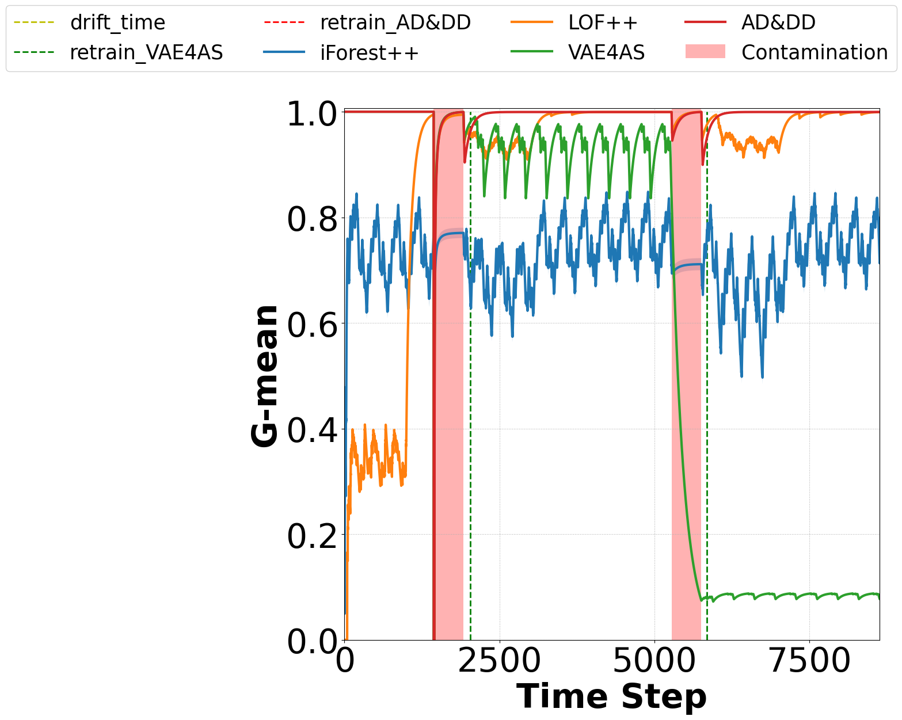} 
  \caption{ZJ\_Sce.1\_N6}
\label{fig:ZJ1_6_com}
 \end{subfigure}%
  \begin{subfigure}{.25\textwidth}
  \centering
  \includegraphics[width=1.0\columnwidth]{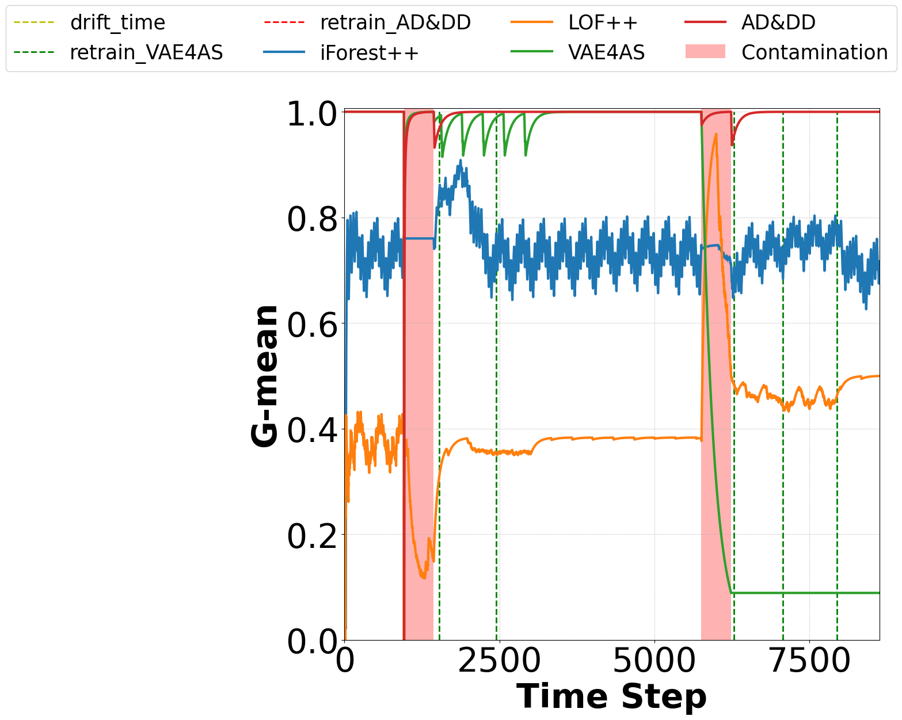} 
  \caption{ZJ\_Sce.3\_N26}
    \label{fig:ZJ3_26_com}
 \end{subfigure}%
\caption{ Performance of iForest++, LOF++, VAE4AS and AD\&DD.}
\label{fig:com}
\end{figure}

\subsection{Contamination localization} 
In this experiment, four scenarios are used, with two for each network. Localization accuracy, as summarized in Table~\ref{tab:loc}, shows that up to 5 nodes, including the contamination source, are localized despite limited sensors. While false negatives are low, false positives are relatively high, likely due to concept drift.

\subsection{Contamination detection performance comparison}

In this section, we compare the detection performance of iForest++, LOF++, VAE4AS, and AD\&DD across four scenarios: N7 in Hanoi Sce.1, N18 in Hanoi Sce.2, N6 in ZJ Sce.1, and N26 in ZJ Sce.3. Contamination periods are highlighted in red, drift periods in yellow dashed lines, and retraining times are marked with red (AD\&DD) and green (VAE4AS) dashed lines.

As shown in Fig.~\ref{fig:com}, VAE4AS performs similarly to AD\&DD in the Hanoi network scenarios but poorly in others, showing limited robustness. While both methods include drift detection, VAE4AS raises false alarms except in Hanoi Sce.2, while AD\&DD accurately detects drift across all cases. iForest++ maintains an average G-mean above 0.7 but does not match AD\&DD's performance. LOF++ shows significant variability, performing well in some scenarios like Hanoi Sce.2 but poorly in others. These results suggest the other methods have limited fault detection potential, so comparative experiments for contamination localization were not conducted with these methods.

\section{conclusions}\label{sec:conclusion}
Contamination in WDNs poses risks to public health, and monitoring chlorine levels helps detect these events as contamination reacts with chlorine. This study simulates arsenite contamination and sensor measurement offsets using two realistic WDNs, proposing AD\&DD, a method combining LSTM-VAE for anomaly detection with a dual-threshold drift detection, requiring no labeled data. Experiments show that AD\&DD outperforms current methods. Future work will explore more complex contamination scenarios.

% \section*{Acknowledgment}

% \Urlmuskip=0mu plus 1mu\relax
\bibliographystyle{IEEEtran}
\bibliography{paper}

\end{document}